\documentclass{article}

\usepackage[preprint]{nips_2019}

\usepackage[utf8]{inputenc} 
\usepackage[T1]{fontenc}    
\usepackage{hyperref}       
\usepackage{url}            
\usepackage{booktabs}       
\usepackage{amsfonts}       
\usepackage{nicefrac}       
\usepackage{microtype}      
\usepackage{graphicx} 
\usepackage{amsmath}
\usepackage{multirow}
\usepackage{xcolor}
\usepackage{subcaption}

\title{ Brain2Char: A Deep Architecture for Decoding Text from Brain Recordings} %

\author{ 
  Pengfei Sun  \\
  Center for Integrative Neuroscience UCSF \\
  \And
  Gopala K. Anumanchipalli \\
   Center for Integrative Neuroscience UCSF \\
   \And
   Edward F. Chang \\
   Center for Integrative Neuroscience UCSF \\
 }

\begin{document}
\maketitle

\begin{abstract}
Decoding language representations directly from the brain can enable new Brain-Computer Interfaces (BCI) for high bandwidth human-human and human-machine communication. Clinically, such technologies can restore communication in people with neurological conditions affecting their ability to speak. In this study, we propose a novel deep network architecture Brain2Char, for directly decoding text (specifically character sequences) from direct brain recordings (called Electrocorticography, ECoG). Brain2Char framework combines state-of-the-art deep learning modules --- 3D Inception layers for multiband spatiotemporal feature extraction from neural data and bidirectional recurrent layers, dilated convolution layers followed by language model weighted beam search to decode character sequences, optimizing a connectionist temporal classification (CTC) loss. Additionally, given the highly non-linear transformations that underlie the conversion of cortical function to character sequences, we perform regularizations on the network's latent representations motivated by insights into cortical encoding of speech production and artifactual aspects specific to ECoG data acquisition. To do this, we impose auxiliary losses on latent representations for articulatory movements, speech acoustics and session specific non-linearities. In 3 participants tested here, Brain2Char achieves 10.6\%, 8.5\% and 7.0\% Word Error Rates (WER) respectively on vocabulary sizes ranging from 1200 to 1900 words. Brain2Char also performs well when 2 participants silently mimed sentences. These results set a new state-of-the-art on decoding text from brain and demonstrate the potential of Brain2Char as a high-performance communication BCI. 
\end{abstract}

Several demonstrations in recent years have shown that it is possible to decode cognitive, linguistic and speech representations directly from the brain through machine learning on neurophysiological imaging datasets. Attempts have been successful in decoding word classes or semantic representations from fMRI data (Mitchell, et al. (2008), Wehbe, et al. (2014), Huth, et al. (2016), Pereira, et al. (2018)). However, since speech communication happens at a much faster rate, accurately decoding cortical function at the rate of fluent speech requires neurophysiological imaging at higher spatial resolution (in the order of millimeters) and temporal resolution (in the order of milliseconds). Among the modalities that offer the best resolution for assaying neural function is Electrocorticography (ECoG, Sejnowski, et al. (2014), Chang, et al. (2015)). ECoG is an invasive neuroimaging technique where a flexible array of electrodes (3 in $\times$ 3 in) is placed directly on the surface of the cortex as part of clinical treatment for intractable epilepsy. Each electrode records the raw voltage potentials at the cortical surface, an aggregate electrical activity of thousands of neurons underneath the contact. This setting provides a unique opportunity to create datasets of parallel neural and behavioral data as participants perform tasks such as listening or speaking naturally. Indeed, several key results have been published in recent literature on inferring speech representations directly from associated ECoG activity, broadly referred to here as Neural Speech Recognition (NSR). 

Prior works on speech/language decoding from ECoG used data either from the auditory or the speech motor cortices. Pasley et al. (2012) and Akbari et al. (2019) report methods for direct reconstruction of external speech stimuli from auditory cortex activations. Martin et al. (2016) use non-linear SVM  to classify auditory stimuli and Moses et al. (2016) use Hidden Markov Modeling to infer continuous phoneme sequences from neural data during listening. Similarly, for neural decoding during speech production, Mugler et al. (2018) have used linear models to classify phonemes and Herff et al. (2015) use HMM/GMM based decoding to achieve a WER of 60\% on a 50 vocabulary task. Recent attempts have focused on decoding audible speech directly from the sensorimotor cortex, including Angrick et al. (2019), using deep convolutional architectures (Wavenet) and Anumanchipalli et al. (2019), using recurrent architectures.

In speech processing applications like automatic speech recognition (ASR) and text-to-speech synthesis (TTS), much progress has been made to achieve near-human performance on standard benchmarks. These include the use of recurrent (Hannun et al. (2014)) and convolutional architectures (Collobert et al. (2016)) towards End-to-End speech recognition minimizing CTC loss, which remain unexplored for decoding text from brain data. Given the demonstration of  naturalistic and continuous speech synthesis directly from the brain, one way to approach the Brain-to-text problem is a 2 stage model where neural data is converted to speech, which can then be converted to text using a state-of-the-art ASR system. Indeed, we propose a strong baseline for the Brain-to-text problem along these lines, that to our knowledge, already achieves the best reported performance on this task. To improve this baseline further, we propose Brain2Char, a deep network architecture that borrows ideas from speech processing, and proposes optimizations appropriate for NSR. Brain2Char implements an End-to-End network that jointly optimizes various sub-problems, like neural feature extraction, optimizing latent representations and session calibration through regularization via auxiliary loss functions. Performance of Brain2Char is quantified on 4 volunteer participants who spoke overtly or silently, and various aspects of decoder design are objectively evaluated.

\vspace{-0.05in}
\section {Neural Speech Recognition from ECoG}
\vspace{-0.05in}
Unsupervised modelling techniques have shown great promise in various speech and language applications, but these methods rely on vast amounts of training data. Since ECoG data is acquired from individual participants in clinical settings, the amount of useful training data is limited in most cases. Moreover, since individual brains differ anatomically, neural data cannot be directly pooled across participants. Hence, conventional speech recognition methods do not work right off-the-shelf for neural data without customization. Ideally, a neural speech recognizer is robust to all known aspects of neural and behavioral variability to model the conversion of one to the other. Previous insights from neurobiology have prescribed the analytic amplitude of the High Gamma frequency band (70-150 Hz) as a robust feature that correlates well with multi-unit spiking (Edwards, et al. (2005), Crone, et al. (2011)), and that there is complementary information in lower frequency ranges. Besides these features, there are aspects specific to ECoG and speech production that need to be modelled in NSR. Here are some factors that contribute to neural variance in ECoG signal, as related to speech production --

\textbf{Neural Basis of Speech Production} :  Neural mechanisms for linguistic planning and execution occur at diverse timescales, and at diverse alignment offsets with respect to the speech signal. Also, different cortical regions encode distinct aspects of speaking. For example, the IFG is linked to motor sequence planning (Flinker et al. (2015)) and the vSMC is linked to the  articulatory aspects in producing speech (Chartier, et al. (2018), Mugler, et al. (2014)). Each electrode location in the vSMC codes a unique kinematic plan, spanning multiple vocal tract articulators and timescales to orchestrate continuous speech articulation. Electrodes in the STG encode spectrotemporal aspects of speech signal (Mesgarani et al. (2014)). This diversity of cortical function and tuning properties contributes to most of the speech related variability in the ECoG signal. It is therefore important for an NSR system to model these representations, as appropriate for a given electrode location and participant behavior (e.g., whether it's a speaking task or a listening task).

\textbf{Intrinsic Neural Variability}: ECoG measures the local field potentials at the cortical surface. Since each electrode contact records from thousands of underlying neurons, the signal at each may not be entirely specific to speech production. Some cortical regions responsible for certain articulatory phonetic sequences may not even be sampled, given a particular electrode coverage, or there may be redundancy across neighbouring electrodes. Furthermore, the intrinsic neural dynamics (Churchland, et al. (2010), Sun, et al. (2019)) mask the true signal causal to producing speech. All of these aspects contribute to a poor signal-to-noise ratio (SNR) in the neural data. It is important for NSR systems to extract meaningful features from across the diverse spatial and temporal scales of the ECoG data.

\textbf{Cross-session variability}: Typical ECoG data is collected in multiple, short sessions spread over a week or so, where electrode leads from the brain are manually connected to a preamplifier before recording neural data for each session. An artifactual aspect of neural recordings is the non-stationarity in the signal across recording sessions. This may be due to different ``baseline" brain states in each session,  or some electrode channels not recording (e.g., sensor losing contact with the cortical surface), or (more rarely) the entire grid slightly shifted on the cortical surface, causing differences across sessions that are unrelated to speech production process itself. An ideal NSR approach subsumes session specific corrections to improve cross session compatibility in neural data.

Since most of these issues cannot be deterministically modelled, it is necessary for neural speech recognition to be realized within an unsupervised framework, jointly optimizing neural feature extraction, latent representation learning, session calibration and text prediction, all within the same model. We now describe Brain2Char, a deep learning architecture that implements such a framework. 
\vspace{-0.1in}
\subsection{Brain2Char Architecture}
\begin{figure*}[h]
     \centering
     \vspace{-0.1in}
     \includegraphics[trim={0 0 0 0.40cm},clip, width=1.0\textwidth]{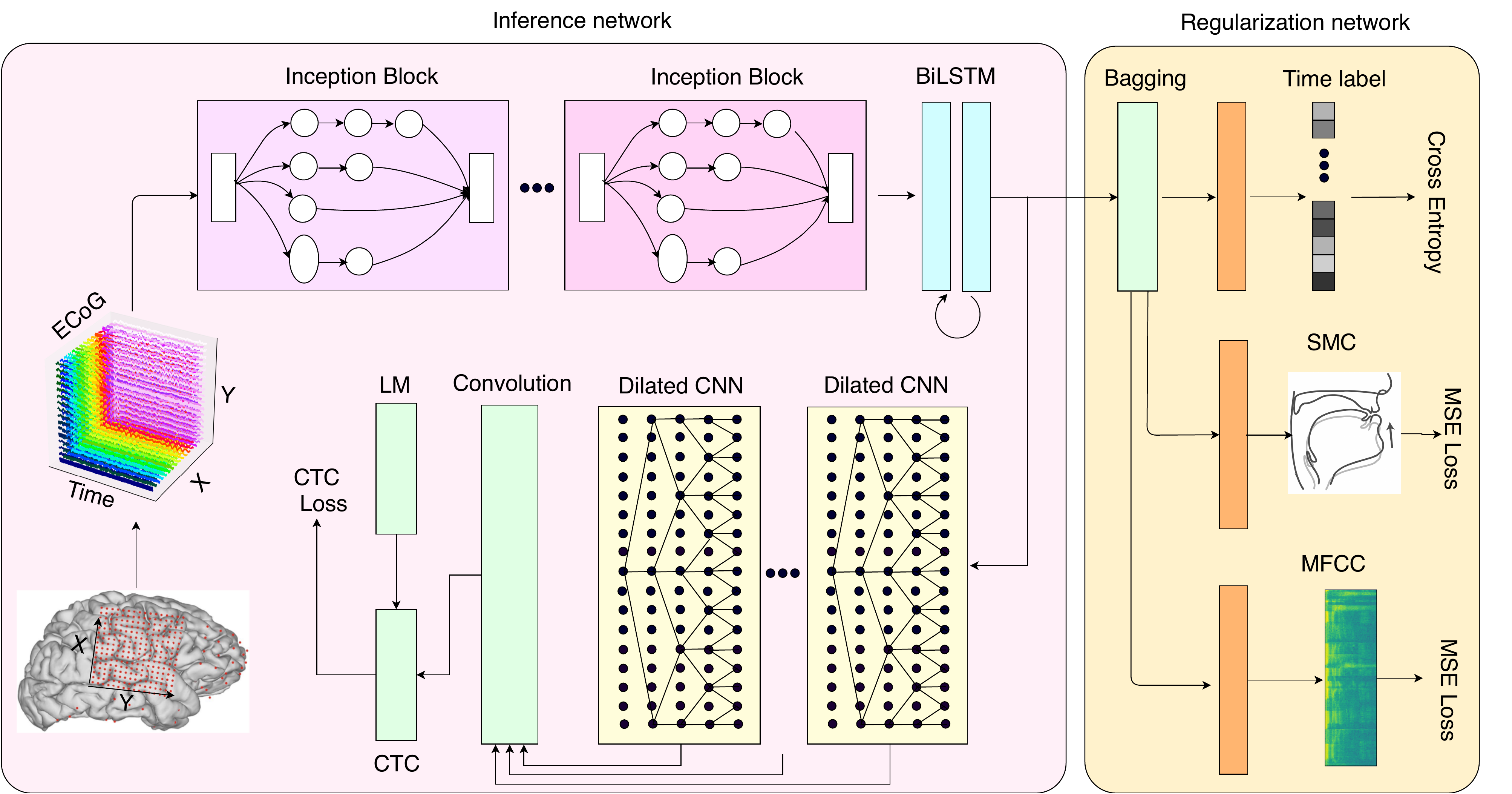}
     \caption{\small Architecture of Brain2Char: Neural data is recorded as participants produce speech. Different filters spanning multiple space, time and frequency dimensions convert recorded potentials into appropriate feature representations. The intermediate features are fed into regularization network and the decoder network. The regularization networks impose MSE losses on regressed speech representations and session embeddings. The decoder implements a sequence learning model that dilated CNNs convert the latent representations to character sequences, using language model weighted BeamSearch.}
     \label{fig:decoder}
     \vspace{-0.1in}
\end{figure*}

\vspace{-0.05in}
We propose a neural speech recognition framework Brain2Char with a modular architecture comprising three parts: the neural feature encoder, the text decoder and the latent representation regularizer. The modular structure is convenient for network optimization, and each sub module can be independently improved based on the general design considerations of NSR systems mentioned in the earlier section. The inference model consists of the encoder and the decoder, and the regularization networks are only used at training time. Figure\ref{fig:decoder} illustrates the architecture of the Brain2Char decoder. 

\textbf{Notation}: Brain2Char optimizes a transformation $\Phi$ to map the recorded neural signals $X=\{x_{1}, x_{2}, \cdots, x_{n}|x_{i} \in \mathbb{R}^{t\times w \times h}\}$ to character sequences $Z=\{z_{1}, z_{2}, \cdots, z_{m}| z_{j} \in \mathbb{R}^{v}\}$. Here the index $t, w, h$ refer to time dimension, the anterior-posterior (width) and dorsal-ventral (height) axes of the ECoG grid, and $v$ refers to the embedding dimension of text vector.  In the encoding phase, encoder $\Phi_{e}$ projects neural inputs $X$ to latent feature space $F$, which in turn will be translated as the outputs $Z$ by decoder $\Phi_{d}$ in the decoding phase. Here capital $F$ with upper index (e.g., $F^{h}$) represents the basis, whereas with the foot index (e.g., $F_{h}$) is vector. 

\subsection{Neural Feature Encoder Network}
 \vspace{-0.05in}
The goal of the encoder is to extract the speech-specific part of the neural signal, robustly accounting for the spatial, temporal and spectral variations within the neural signals. In Brain2Char, the 3-dimensional ECoG signals $X$ are fed into the encoder stacked by network modules similar to inception nets (Szegedy et al. (2017)). A single inception layer parallelly employs several sub-networks to extract features at different resolutions, by choosing a set of hyperparameters specifying the number of channels, kernel size and number of stacked layers. This design is capable of extracting features at various space-time scales within single inception module, and aggregate these multi-resolution features. Therefore, the next stage can access robust features at different resolutions simultaneously. 

The neural feature encoder $\Phi_{e}$ consists of several layers of grouped convolutional neural networks (CNN). For each single layer, a set of convolutional filters $\{\mathcal{W}_{1}, \mathcal{W}_{2}, \cdots\}$ are used in multiple sub-networks, where $\mathcal{W}_{i} \in  \mathbb{R}^{c_{in} \times k_{t}\times k_{w} \times k_{h}\times c_{out}}$. For the weight tensor $\mathcal{W}_{i}$, the spatial support of each kernel filter $\kappa_{i}$ is $k_{t}\times k_{w} \times k_{h}$. There are $c_{in}$ input channels and $c_{out}$ output feature maps. with the input tensor $\Lambda \in \mathbb{R}^{t_{in} \times w_{in}\times h_{in}\times c_{in}}$, the convolution between kernel function $\kappa_{i}$ and the input tensor $\Lambda$ can be translated as $\mathcal{F}(\kappa_{i} * \Lambda) = \mathcal{F}(\kappa_{i})\odot \mathcal{F}(\Lambda)$, where $\mathcal{F}$ represents Fourier transform, $*$ and $\odot$ refer to convolution and Hadamard product, respectively. By utilizing different kernel sizes $k_{t}, k_{w}, k_{h}$, each $\mathcal{W}_{i}$ can extract components in certain frequency bands by imposing a group of kernel filters. Additionally, to ensure the extracted features are densely sampled at diverse resolutions, each sub-networks $p$ in $j$th layer may use $l$ stacked layers that employ a series of kernels $\kappa$ with different sizes. Therefore, the output of sub-networks can be written as
\begin{equation}
    F_{\{j,p | \kappa\}}=\underbrace{\mathcal{W}_{i}(\mathcal{W}_{q}(\cdots(}_{l}\Lambda))),
\end{equation}
where $F_{\{j, p|\kappa\}}$ refers to the features obtained at the $p$th sub-networks in $j$th layer of encoder by using kernel set $\{\kappa\}$. After multi-scale feature extractor, two Bi-LSTM layers are stacked to strengthen the sequence correlation learning within the latent representations. In the proposed framework, the output of the Bi-LSTM layers are referred as the latent feature representation $F_{h}$. 
\vspace{-0.05in}
\subsection{Regularization network}
\vspace{-0.05in}
To implicitly enforce a meaningful latent representation in the neural encoder, the regularization branch performs simple feed-forward transformations of the latent features of the encoder to account for known aspects of neural signal variance discussed earlier.

{\bf Session Calibration}: To reduce the variabilty of neural signals $X$ due to session specific artifacts, calibration across sessions is done to pool sessions in terms of their relative similarity to the earlier sessions. Since changes across sessions cannot be modelled systematically, in this study we introduce an implicit calibration on the encoded feature $F$ to reduce intrinsic session-dependent variation. In other words, each latent feature $F_{t, t\in T_{i}}$ within the time duration $T_{i}$ can be assigned session labels $S_{T_{i}}$, which are indexed by continuous valued numbers. Instead of using one-hot vector to represent $S_{T_{i}}$, we learned the time embedding vector $Q_{T_{i}}$ in a \emph{\bf skip-gram} fashion (Mikolov et al. (2013)). Similar to word embedding, $Q_{T_{i}}$  describes the correlation among $S_{T}$ based on their temporal positions. A regression layer $M(\cdot)$ would map the dynamic latent sequence $F_{h}$ to the embedding $Q_{T_{i}}$, and $\mathcal{L}=\|M(F_{h})- Q_{t}\|^{2}$ is used as the cost function of time regularization.

{\bf Speech-specific latent representation}: The basis space $F^{x}$ representing neural signals $X$ exist in a higher dimensional space compared to the basis space $F^{z}$ of text $Z$. In Brain2Char architecture, the encoder projects the neural data $X$ into the reduced latent space $F^{h}$ and the decoder expands the latent feature $F_{h}$ to span the basis of targets $Z$. If trained without regularization on the output of $\Phi_{e}$, the encoder $\Phi_{e}$ explores quite a large space searching for a manifold where $F^{z} \subset F^{h}$. To reduce the search space, the language basis $F^{z}$ can be directly used as the regression target of $\Phi_{e}$. However, limited by the data, complete basis $F^{z}$ is not obtainable. As the baseline we proposed, other accessible feature representation $F^{z^{\prime}}$, such as MFCC, can be used as the regression targets of $\Phi_{e}$. Directly employing $F^{z^{\prime}} \subset F^{z}$ as the output of encoder may over penalize the ECoG basis $F^{x}$. Therefore, a more practical approach is to keep $F^{h}$ as the intermediate layer and apply empirical feature space $F^{z^{\prime}}$ as the regularization. It retrains the pattern variance of $F^{x}$, whereas utilizes MFCC from speech acoustics or articulatory kinematic movement to inform intrinsic feature space of $F^{h}$. That is $\{F^{z}\setminus F^{z^{\prime}}\} \subset F_{h}$. Generally, the regularization features $F^{z^{\prime}}$ can be any type of features correlated with or generated from speech or text $Z$ associated with productions. For instance, by applying autoencoder on speech acoustics, Akbari et al. (2018) derive a low dimensional basis  utilized as the regularization component in auditory speech reconstruction. Assuming a set of feature basis $\{F^{s_{1}},F^{s_{2}},\cdots\}$, we can either utilize these feature basis independently, or in a more ensemble fashion, where a joint feature basis is learned. Here the index $s_{i}$ refers to $i$th category features (e.g., MFCCs). Based on the obtained feature vectors $\{F_{s_{i}}\}$, two types of regularizations are described as:
\begin{equation}
\begin{split}
    \Psi(F_{h}, F_{s_{1}}, \cdots, F_{s_{i}}) &=  \sum_{i}\alpha_{i} \|F_{h}-F_{s_{i}}\| \\
    \Psi(F_{h}, F_{s_{1}}, \cdots, F_{s_{i}}) &=  \|F_{h}-\Omega(F_{s_{1}}, \cdots, F_{s_{i}})\|, 
\end{split}
\label{feature_}
\end{equation}
where $\Omega$ represents projection operation, and $\Psi$ is cost function. By using $\Omega$, a new ensemble feature basis is incorporated to modulate the latent representation. Physiologically generative representations or close features derived  from speech acoustics make better targets for regularization. 

\vspace{-0.05in}
\subsection{Text Decoder Network}
\vspace{-0.05in}
At the core, translating the latent feature $F_{h}$ to character sequences is a sequence decoding task. In other words, any state-of-the-art sequence translation system can be adapted to the text decoder network. In the context of our application, different temporal scale of latent features $F_{h}$ must be used as appropriate, as the relationship between text and speech is temporally imprecise. Brain2Char model employs three layers of dilated CNNs to process the long-short term correlations, which could be more resistant to noise components (in comparison to sequence to sequence decoding, that decouples the features by locally transferring the state, and obviously is more vulnerable to background noise interference). In each dilated layer, 5 sub-layers as shown in fig.\ref{fig:decoder} are applied to learn the sequence correlations at various scales. The layer-wise residual connections ensure the features at different scale are processed simultaneously. Since the exact alignment of latent representations is hard to obtain (i.e., onset and offset of neural data corresponding to characters), on top of the dilated CNN, CTC is incorporated with 4-gram language model as the text decoder network. Together with feature regularization and the explicit language model used for beam search, the cost function for the overall Brain2Char decoder is
\begin{equation}
    \mathcal{L}  = \alpha_{0}\mathcal{L}_{I} +\sum_{s_{i}}\alpha_{s_{i}}\mathcal{L}_{s_{i}}, 
\label{LossFun}
\end{equation}
where the first component $\mathcal{L}_{I}$ refers to the loss of the inference networks, and the second component is the summation of the loss of regularization networks.

\textbf{Implementation} We implemented our method using Tensorflow. In terms of model training, we use a cyclic learning rate with maximal value 0.005 and minimal value 0.0001. Linear decay coefficients are applied to the weight coefficients of regularization components. The batch size is 50 at the sentence level for training. The feature dimension of MFCC and AKT are 26 and 33, respectively. For 3D inception module, the kernel sizes $k_{t}, k_{w}, k_{h}$ along each axle are selected from $\{1, 3, 5, 7\}$. The dimension reduction is achieved by using 2 step strides in inception modules. For dilated CNN, the filter size is fixed as 11, and the dilated ratios for five sub-layer are $[1, 2, 4, 8, 16]$. The convlutional layer after the dialted CNN uses kernel size 1, and the output channel is the dimension of character vocabulary. Two BiLSTM layers are set at 0.5 dropout rate, and the output of dilated CNN is set 0.15 dropout rate. To ensure robust sequence learning, the onset and offset of neural features are randomly jittered about a time window aligned to speech boundaries. A 4-gram language model incorporated with beam search is be applied to the outputs of CTC. KenLM (Heafield et al. (2013)) is used to train the word language models of the speech corpus that is used for speaking task. Additionally, the pre-trained \emph{LibriSpeech} language model in \emph{DeepSpeech} (Hannun et al. (2014)) is used as a baseline language model. The weighting coefficient of language model is set to 1.5. 
\vspace{-0.05in}
\section{Experimental Results}
\vspace{-0.05in}
 \textbf{Data}: In this study, data from four participants $P1, P2, P3$, and $P4$ is used. These volunteer participants read  prompted sentences on a screen while their speech and ECoG data were synchronously recorded. The sentences were derived from MOCHA-TIMIT corpus (a 1900 word vocabulary task of 460 independent sentences, for participants $P1$ and $P2$) and a limited domain dataset of verbal descriptions of 3 pictures (a 400 words vocabulary on controlled description and 1200 vocabulary on free-style interview task for participants $P3$ and $P4$). The total data collected across participants varied between 120 minutes to 200 minutes. Participants $P3$ and $P4$ also silently mimed a subset of 20 sentences without overt vocalization. The recordings were made in several one hour long sessions, over a week or more while participants were implanted with grids for clinical monitoring for seizure localization. Specifically, the neural data of subject $P1$, $P2$ and $P4$ are recorded with 16 $\times$ 16 electrode grid covering the ventral sensorimotor cortex (vSMC), inferior frontal gyrus(IFG) and superior temporal gyrus(STG), only subject $P3$ was recorded with 16 $\times$ 8 electrode grid covering only the dorsal half of the vSMC. All neural data was preprocessed to reject artifacts and extract the analytic amplitude in the High Gamma frequency band(70-150Hz) and low frequency component (0-40 Hz) z-scored appropriately. For the speech data collected, acoustic-to-articulatory inversion is performed to estimate the articulatory kinematic trajectories (AKT), and Mel Frequency Cepstral Coefficients (MFCC) are extracted. All data were synchronously sampled at 200 Hz. 
 
 \textbf{Baseline}: We designed a baseline ECoG-to-text system inspired by previous demonstrations of speech synthesis from the ECoG data (Angrick, et al. (2019) and Anumanchipalli, et al. (2019)), and off-the-shelf ASR systems. We used \emph{DeepSpeech}, a pretrained state-of-the-art ASR system to decode text from the acoustic features, which in turn are estimated from ECoG features using a neural feature encoder $\Phi_{e}$, implemented as a 2 BiLSTM layers. Since neural speech synthesis does not yield perfectly intelligible speech, directly decoding neurally synthesized MFCCs through \emph{DeepSpeech} resulted in poor performance (80\% WER). To improve this, the baseline neural feature encoder $\Phi_{e}$ is optimized for the \emph{DeepSpeech} network. The neural feature encoder is stacked on top of a pre-trained \emph{DeepSpeech} network and trained on parallel ECoG, MFCC and Text data. The joint network is optimizes a CTC loss, and an optimally weighted auxiliary mean-squared error (MSE) loss on the MFCCs. We found it beneficial to `freeze' the pre-trained layers of \emph{DeepSpeech}, and only allowing the neural encoder layers to be learnt, while still optimizing the joint loss. 

\begin{figure*}[!h]
     \centering
     \includegraphics[trim={0 0 0 0.60cm},clip, width=0.45\textwidth]{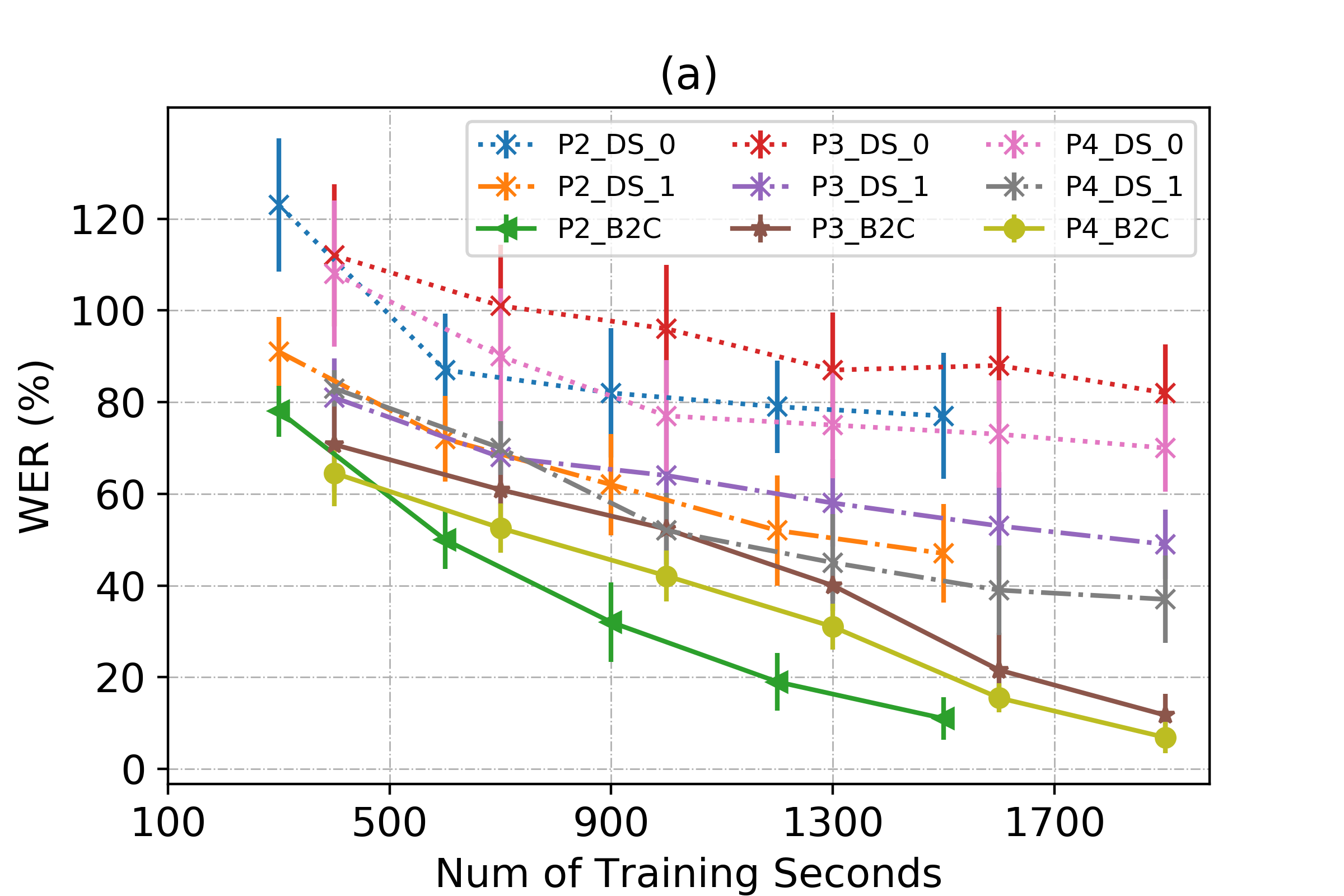}
      \includegraphics[trim={0 0 0 0.60cm},clip, width=0.45\textwidth]{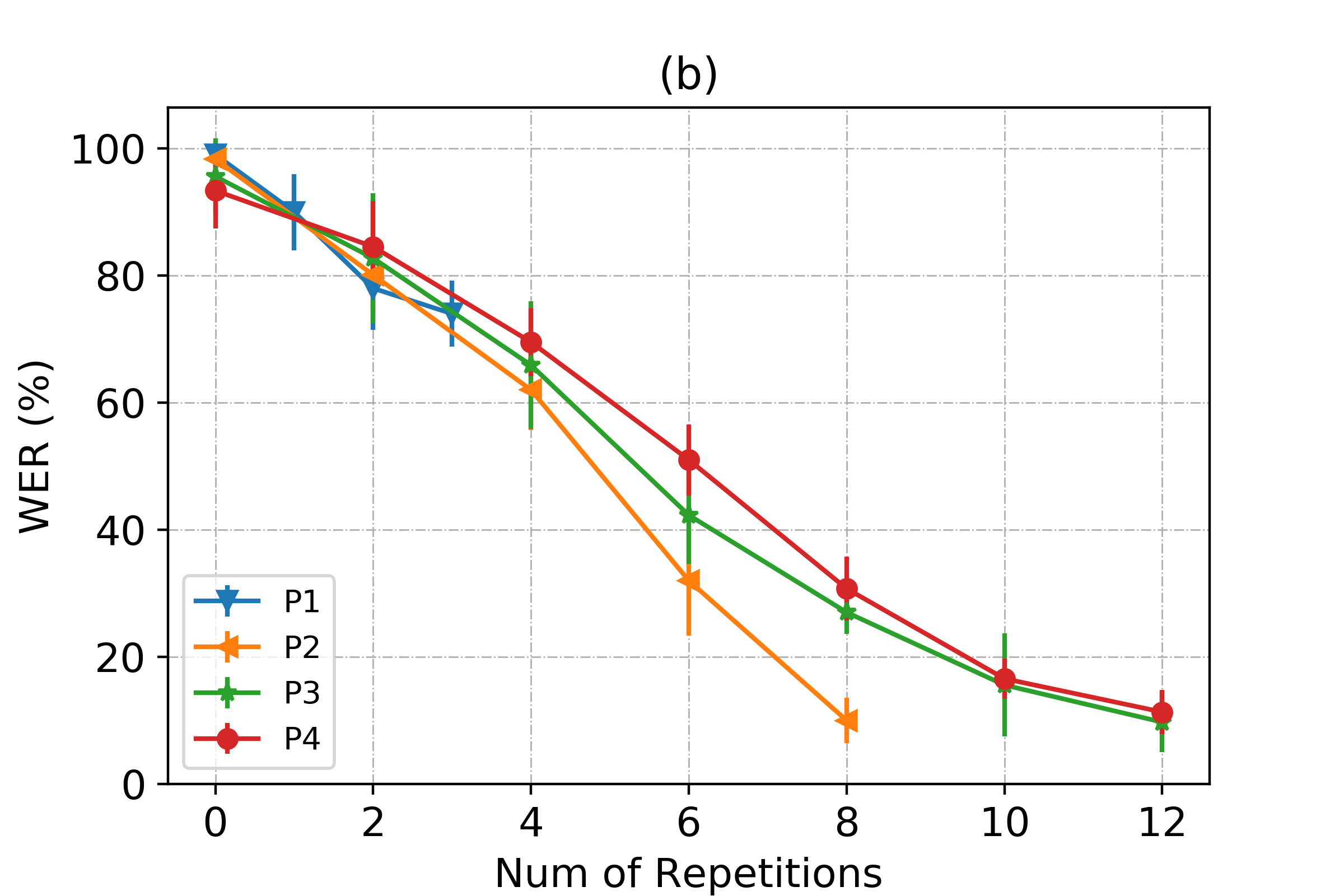} \\
      \includegraphics[trim={0 0 0 0.30cm},clip, width=0.45\textwidth]{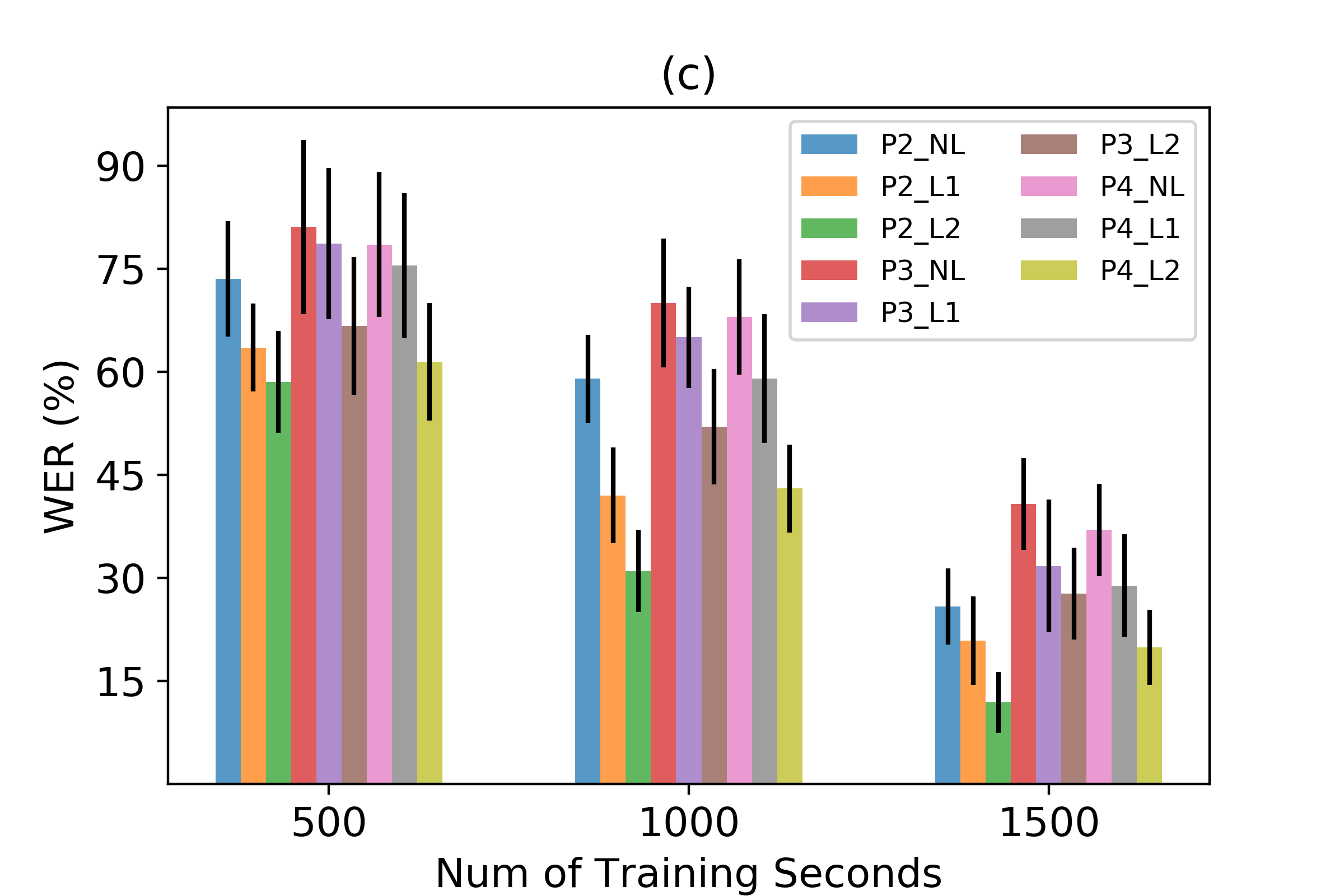}
      \includegraphics[trim={0 0 0 0.30cm},clip, width=0.45\textwidth]{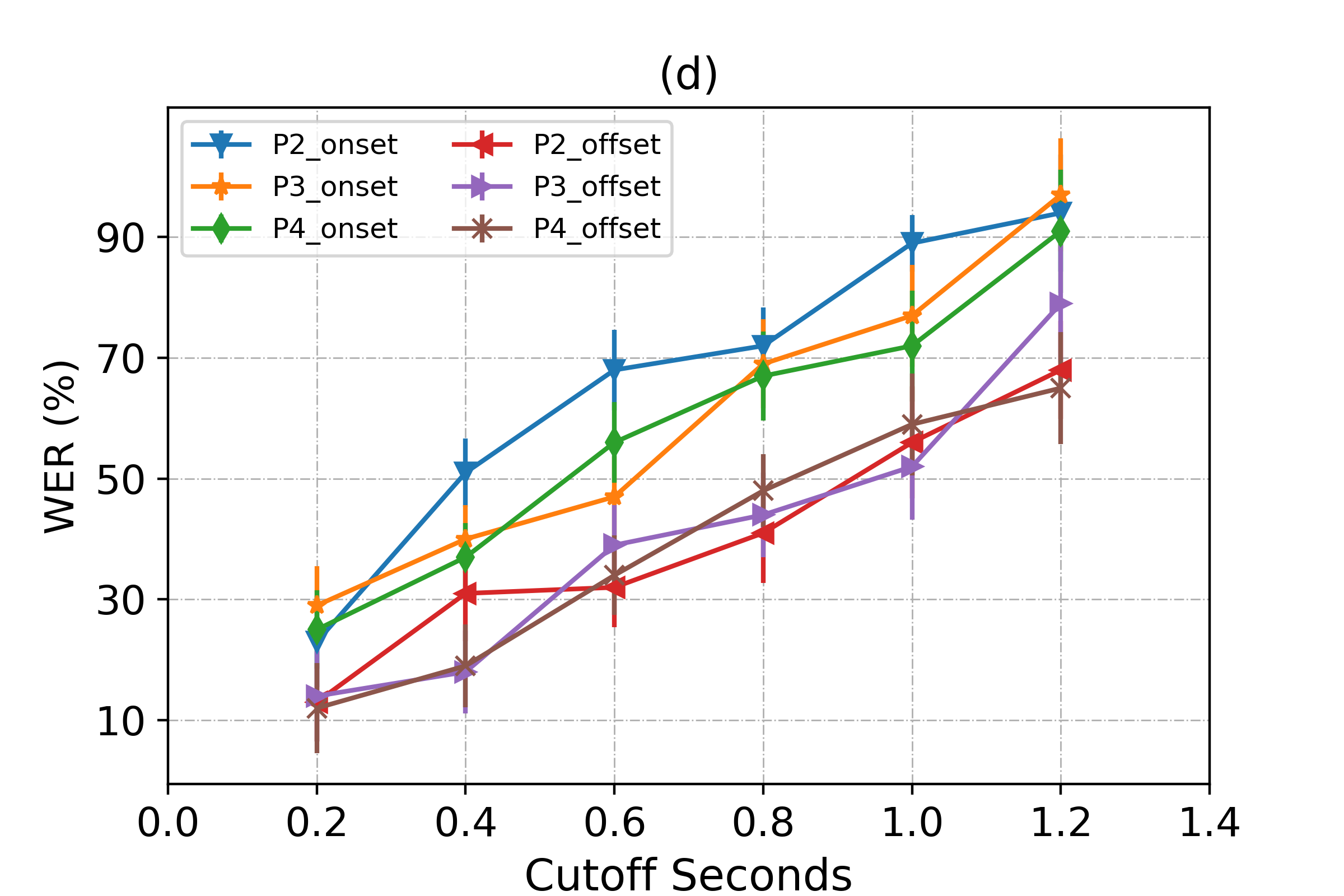}
     \caption{\small Performance evaluation of Brain2Char (a) WER as a function of increasing amount of training data for baselines and Brain2Char. (b) The effect on WER of sentences against the numer of trials for a given sentence provided in training. (c) Contribution of various language models. (d) Performance on partial neural data with cutoff on either onset or offset.} 
     \label{training_time}
     \vspace{-0.2in}
 \end{figure*}

\textbf{Quantitative results}. 
We conducted a quantitative evaluation of the baselines and the proposed Brain2Char architecture. Fig\ref{training_time}(a) compares the performance of various systems with increasing amounts of training data. Systems indexed DS$_{0}$ are a 2-stage implementation of speech synthesis from the brain using neural feature encoder $\Phi_{e}$,  followed by pretrained \emph{DeepSpeech} to decode text. The baseline DS$_{1}$ refers to the joint network where $\Phi_{e}$ is modulated by \emph{DeepSpeech}, trained jointly optimizing the CTC loss on characters and MSE loss on intermediate MFCCs. Across 3 patients, the joint optimized DS$_1$ systems show consistent gains of around 30\% in WER over DS$_0$. This suggests that acoustic features alone are not sufficient to code neural representations and customizing the intermediate representations in $\Phi_{e}$ is critical. Fig\ref{training_time}(a) also shows the performance of the proposed Brain2Char networks (indexed B2C) against these baselines. In all cases, we see significant gains of an additional 30\% in WER, and the performance trends with time also suggests that the architecture makes optimal utilization of the available data compared to the baseline (larger slope in WER gain with more data, whereas the baselines seem to plateau). 

Since the training data from each subject is not an exhaustive representation of all phonetic aspects of English language, we wanted to see how important it is to have observed similar phonetic contexts in the training data. This analysis was done by varying the number of repetitions of the same sentence in several training runs of Brain2Char networks. Fig\ref{training_time}(b) shows the WER as a trend against the number of prior repeats of testing sentences in training data (in other trials than those in test). It is clear that having several prior example trials of the sentence helps in future decoding of the same sentence at inference time. 

Fig\ref{training_time}(c) quantifies the contribution of language models towards Brain2Char Performance. For different amounts of training data and  for 3 participants,  3 different language modelling conditions at inference -- decoding with no language model (indexed \emph{NL}), the default language model in $DeepSpeech$ trained on \emph{librispeech}, a general purpose character-level language model of English ( $\_$\emph{L1}), and a task specific language model created using all training data from the task($\_$\emph{L2}). It can be seen that language models help improve decoding performance, and task dependent language models help further. In all, Brain2Char achieves roughly 10$\%$ to 25$\%$ WER improvement across three subjects, and the performance improvements against \emph{librispeech} based language model range from 3\% up to 15\%. It is to be noted that the performance is still respectable in the $\_$\emph{NL} conditions, possibly due to implicit language modeling in the text decoder. This may sometimes lead to overfitting if the decoder is simply memorizing the task language independently of the neural data. To test this, we ran inference on trials that were randomly cut off, either at the start or the end. Fig\ref{training_time}(d) shows the error rates as a function of amount of signal cut off (in seconds, either at the start, or at the end of a trial, indexed $\_$\emph{onset} and $\_$\emph{offset}). The results confirm that Brain2Char is sensitive to the length of the trial in time, without completely memorizing entire sentences. It also seems that onset cutoff is worse than offset cutoff. 
 
 The offset cutoff condition in Fig\ref{training_time}(d) also shows that Brain2Char is capable of synchronous, incremental decoding (instead of waiting for  whole sentence length neural data inputs), which is a critical desirable of a real time communication BCI.  Table~\ref{sequence_decoding} shows the performance of Brain2Char on 2 example sentences as data is provided in increments of 0.2 seconds at inference time. Note that the decoded sentences are shorter in length, as would be expected for shorter time windows of neural inputs, and the errors are typically in the last word(s) that may be cutoff midword.
\begin{table*}[h]
\vspace{-0.05in}
  \caption{Incremental decoding demonstration}
  \label{sequence_decoding}
  \centering
  \begin{tabular}{lll}
    \toprule
           &if only the mother could pay attention to her children  &i think their water bill will be high     \\
    \midrule
    0.2s      & \textcolor{red}{his}       & \textcolor{red}{in}     \\
    0.4s      & if only \textcolor{red}{to} & i think \textcolor{red}{the}    \\
    0.6s      & if only the mother \textcolor{red}{a}       &i think their water   \\
    0.8s     & if only the mother could pay \textcolor{red}{atendion}  &  i think their water bill \\
    1.0s     & if only the mother could pay attention \textcolor{red}{with}        & i think their water bill will be     \\
    1.2s   &if only the mother could pay attention to her children          &i think their water bill will be high      \\
    \bottomrule
  \end{tabular}
  \vspace{-0.05in}
\end{table*}

\begin{figure*}[!h]
     \vspace{-0.10in}
     \centering
        \includegraphics[trim={0 0 0 0.60cm},clip, width=0.45\textwidth]{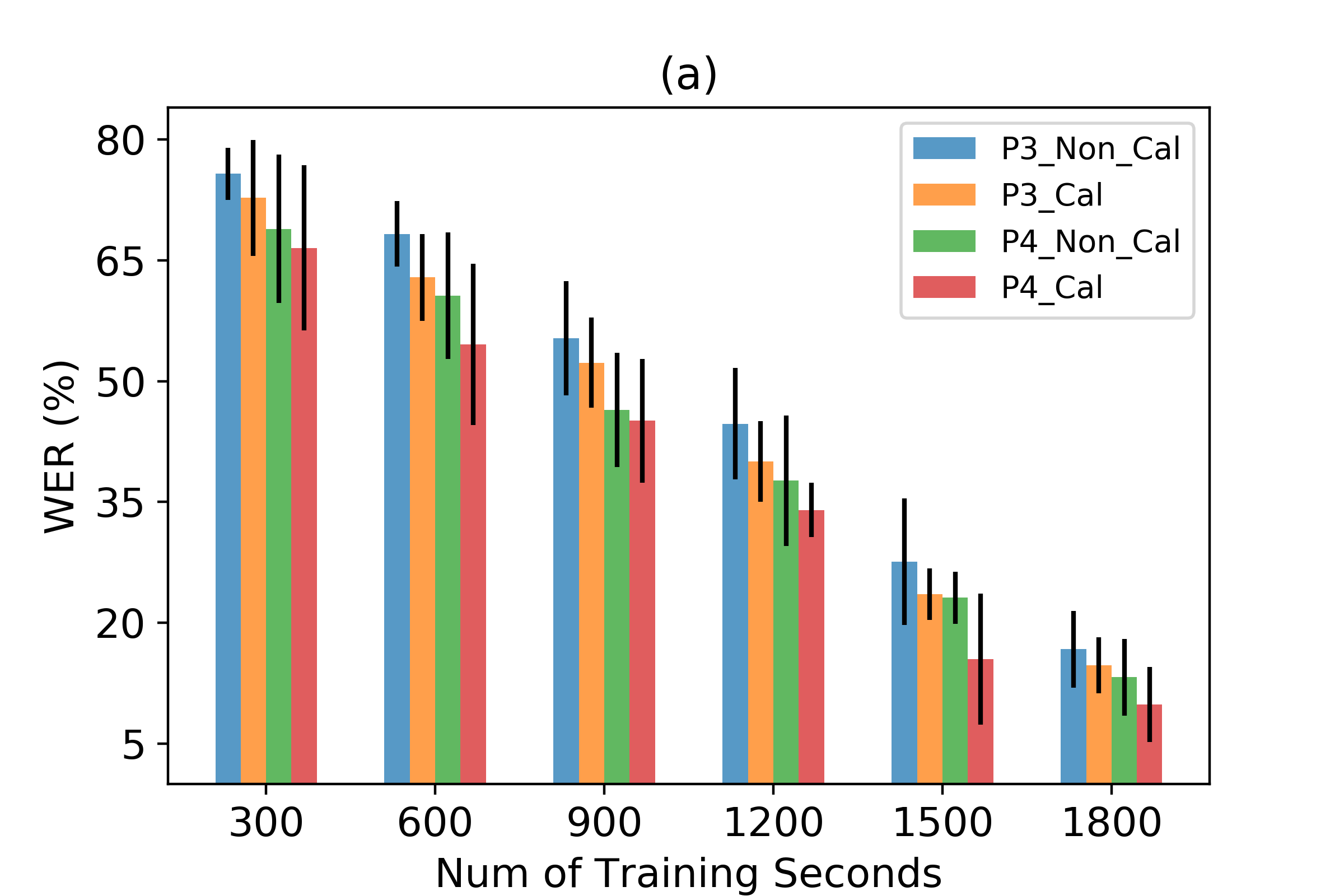}
        \includegraphics[trim={0 0 0 0.60cm},clip,width=0.45\textwidth]{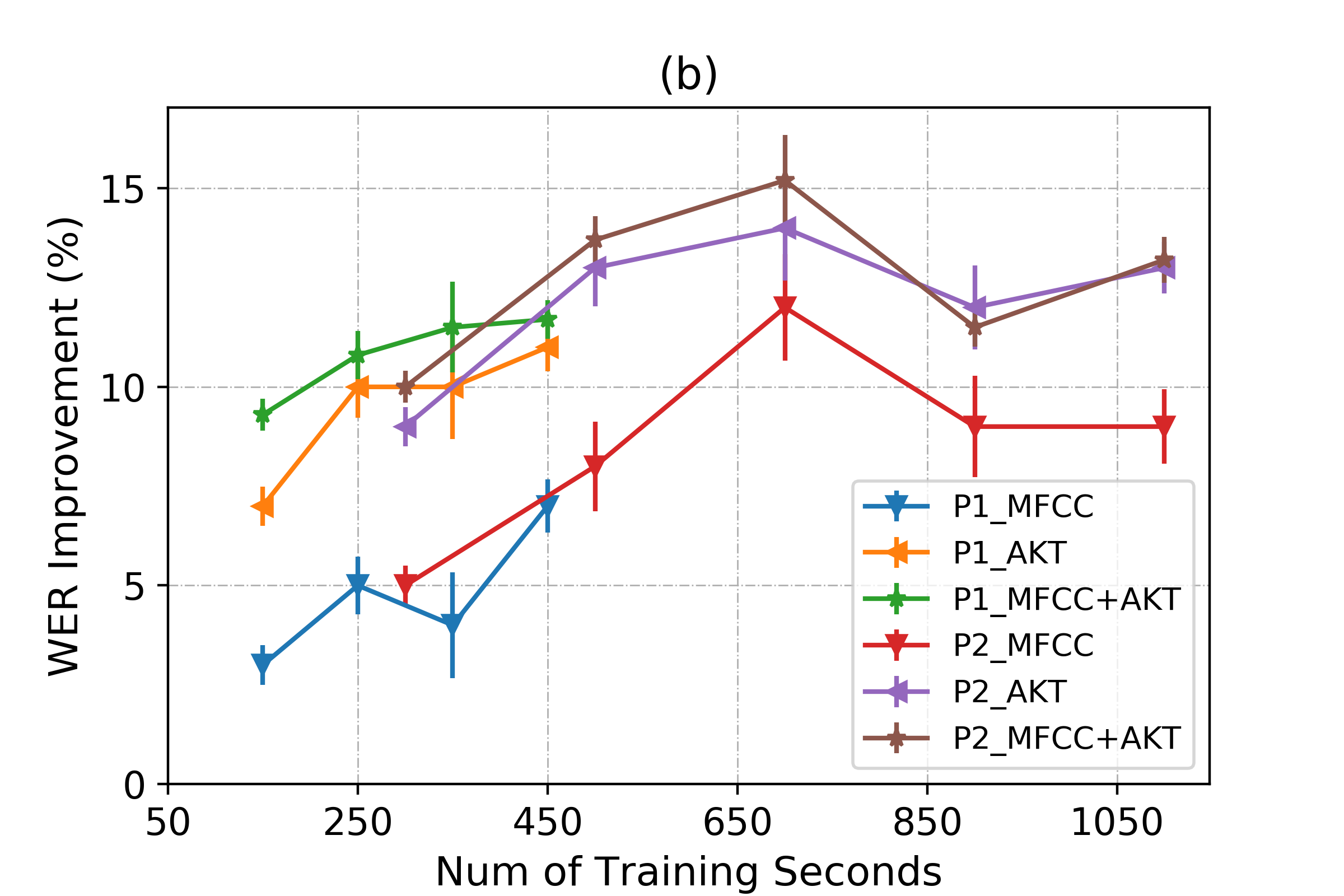}
     \caption{\small Importance of  Regularization factors in Brain2Char. (a) Effect of session calibration on two subjects. (b) WER gains by imposing physiological features targets (e.g., MFCC and AKT)}
     \label{calibration}
     \vspace{-0.1in}
 \end{figure*}

\textbf{Importance of Regularization}: One of the salient features of Brain2Char framework is the regularization branch that implicitly enforces a meaningful and robust latent representation in the neural encoder. To quantify the effect of various regularization factors used, we trained several systems each with different regularization strategies. Firstly, to study the effect of the session embedding regularization (calibration), we trained comparable systems where only in the calibrated condition, the latent representation $F_{h}$ regresses to the attached time embedding. The imposed time constraints on $F_{h}$ reduce cross-session neural variability, and the results in fig.\ref{calibration}(a) confirm this trend, across increasing amounts of training data. In general, the session calibration enhanced the performance by about 4$\%$ against non-calibrated approach.

The second regularization we evaluated was that of the latent speech representation in $F_{h}$. We built variants of Brain2Char systems, where $F_{h}$ was unconstrained (no regularization), and added regularization branches from $F_{h}$ to either i) acoustic features (MFCC), ii) articulatory kinematic features (AKT) and ii) MFCC + AKT. We observed improvements in all these cases compared to the case where no regularization was performed. Fig.\ref{calibration}(b) summarizes these effects in terms of WER improvement from unregularized Brain2Char system. While, all speech representations result in positive gains, articulatory representation are significantly better regularization factors than the spectral MFCC representations. There best improvements were obtained using both representations (MFCC+AKT achieving a 15\% absolute improvement in $P2$), as neural signals may explain some acoustic variations, complementary to the articulatory features. These results indicate that implicitly enforcing physiological aspects in latent representations heavily contribute to explaining the neural speech variance, that cannot otherwise be learnt in an unsupervised fashion, given these smaller scale datasets. The benefits of the articulatory representations are also consistent with earlier studies about neural encoding in the vSMC (Chartier, et al. (2018)).

One confound in neural speech recognition is the possibility of decoding from the auditory regions during the participants' self-perception of their own speech. While, Fig.\ref{calibration}(b) confirms that articulation is the primary driver of the performance, we tested this further by running inference on trials where participants only silently mimed sentences without overtly vocalizing. The decoding performance was still acceptable,  a  40\% WER over 20 tested mimed trials on participant $P3$ and for  $P4$, 67\% WER on 20 mimed trials. These results show that the decoder is not relying on auditory feedback of hearing oneself.  Some selected mimed decoding results are summarized in Table.\ref{mime_decoding}. Note that the errors are often phonetically close confusions (decoded "helping" as ``hoping"; ``sink'' as ``sing" etc). Performance in the silent speaking condition also shows the potential Brain2Char as a silent communication BCI.
\begin{table*}[!h]
  \caption{Mimed speech decoding}
  \label{mime_decoding}
  \centering
  \begin{tabular}{lll}
    \toprule
        &Ground Truth  & Decoded             \\
    \midrule
    \multirow{3}*{P3}  & there is chaos in the kitchen           & \it there is chaos in the \textcolor{red}{catch}   \\
                       &the stool is tipping over    & \it  the stool is tipping over \textcolor{red}{the and}  \\
                       & his sister is helping him steal a cookie &\it his sister is \textcolor{red}{hoping} steal a cookie \\ \\
     \multirow{3}{*}{P4}  & the little girl is giggling   & \it little girl is giggling   \\
                          & water is overflowing from the sink     & \it water is overflowing from the \textcolor{red}{sing}   \\
                          & the woman is holding a broom   & \it the woman \textcolor{red}{the room}  \\
    \bottomrule
  \end{tabular}
  \vspace{-0.20in}
\end{table*}

\section{Conclusion}
We propose Brain2Char, a neural network architecture that converts Brain recordings to text. Brain2Char utilizes multi-scale grouped CNN filters to extract neural signals from ECoG data, employs physiological and artifactual regularization schemes on latent representations, and decodes character sequences optimizing a CTC loss. The jointly optimized Brain2Char model makes optimial utilization of available data and sets a new state-of-the-art performance on decoding text from ECoG recordings. This holds both in terms of vocabulary sizes and performance metrics compared to earlier studies. Furthermore, Brain2Char is amenable to incremental, real-time decoding and performs reasonably well on decoding silently mimed speech, only using brain data. These results demonstrate that Brain2Char is a promising candidate for a communication BCI.


\section*{References}
\small
[1] Mitchell, T. M., Shinkareva, S. V., Carlson, A., Chang, K. M., Malave, V. L., Mason, R. A., \ \& Just, M. A. \ (2008). Predicting human brain activity associated with the meanings of nouns. {\it science}, {\bf 320}(5880): 1191-1195.

[2] Wehbe, L., Murphy, B., Talukdar, P., Fyshe, A., Ramdas, A., \ \& Mitchell, T. \ (2014). Simultaneously uncovering the patterns of brain regions involved in different story reading subprocesses. {\it PloS one}, {\bf 9}(11): e112575

[3] Huth, A. G., de Heer, W. A., Griffiths, T. L., Theunissen, F. E., \ \& Gallant, J. L. \ (2016). Natural speech reveals the semantic maps that tile human cerebral cortex. {\it Nature}, {\bf 532}(7600): 453.

[4] Pereira, F., Lou, B., Pritchett, B., Ritter, S., Gershman, S. J., Kanwisher, N., Botvinick, M. \ \& Fedorenko, E. \ (2018). Toward a universal decoder of linguistic meaning from brain activation. {\it Nature communications}, {\bf 9}(1): 963.

[5] Sejnowski, T. J., Churchland, P. S., \ \& Movshon, J. A. \ (2014). Putting big data to good use in neuroscience. {\it Nature neuroscience}, {\bf 17}(11): 1440.

[6] Chang, E. F., Rieger, J. W., Johnson, K., Berger, M. S., Barbaro, N. M., \ \& Knight, R. T. \ (2010). Categorical speech representation in human superior temporal gyrus. {\it Nature neuroscience}, {\bf 13}(11): 1428.

[7] Chang, E.F. \ (2015). Towards large-Scale, human-Based, mesoscopic neurotechnologies. {\it Neuron}, {\bf 86}: 68-78.

[8] Pasley, B. N., David, S. V., Mesgarani, N., Flinker, A., Shamma, S. A., Crone, N. E., Knight, R. T., \ \& Chang, E. F. \ (2012). Reconstructing speech from human auditory cortex. {\it PLoS biology}, {\bf 10}(1): e1001251.

[9] Akbari, H., Khalighinejad, B., Herrero, J. L., Mehta, A. D., \ \& Mesgarani, N. \ (2019). Towards reconstructing intelligible speech from the human auditory cortex. {\it Scientific reports}, {\bf 9}(1): 874.

[10] Martin, S., Brunner, P., Iturrate, I., Millán, J. D. R., Schalk, G., Knight, R. T., \ \& Pasley, B. N. \ (2016). Word pair classification during imagined speech using direct brain recordings. {\it Scientific reports}, {\bf 6}: 25803.

[11] Moses, D. A., Mesgarani, N., Leonard, M. K., \ \& Chang, E. F. \ (2016). Neural speech recognition: continuous phoneme decoding using spatiotemporal representations of human cortical activity. {\it Journal of neural engineering}, {\bf 13}(5): 056004.

[12] Mugler, E. M., Patton, J. L., Flint, R. D., Wright, Z. A., Schuele, S. U., Rosenow, J., Shih, J. J., Krusienski, D. J., \ \& Slutzky, M. W. \ (2014). Direct classification of all American English phonemes using signals from functional speech motor cortex. {\it Journal of neural engineering}, {\bf 11}(3): 035015.

[13] Herff, C., Heger, D., De Pesters, A., Telaar, D., Brunner, P., Schalk, G., \ \& Schultz, T. \ (2015). Brain-to-text: decoding spoken phrases from phone representations in the brain. {\it Frontiers in neuroscience}, {\bf 9}: 217.

[14] Angrick, M., Herff, C.  Mugler, E., Tate, M. C., Slutzky, M. W., Krusienski, D. J. \ \& Schultz, T.\ (2019). Speech synthesis from ECoG using densely connected 3D convolutional neural networks. {\it Journal of neural engineering}, {\bf 16}, 036019.

[15] Anumanchipalli, G. K., Chartier, J., \ \& Chang, E. F. \ (2019). Speech synthesis from neural decoding of spoken sentences. {\it Nature}, {\bf 568}: 493-498.

[16] Hannun, A., Case, C., Casper, J., Catanzaro, B., Diamos, G., Elsen, E., Prenger, R., Satheesh, S.,  Sengupta, S., Coates, A., \ \& Ng, A. Y. \ (2014). Deep speech: Scaling up end-to-end speech recognition. {\it arXiv preprint} arXiv:1412.5567.

[17] Collobert, R., Puhrsch, C., \ \& Synnaeve, G. \ (2016). Wav2letter: an end-to-end convnet-based speech recognition system. {\it arXiv preprint} arXiv:1609.03193.

[18] Edwards, E., Soltani, M., Deouell, L. Y., Berger, M. S., \ \& Knight, R. T. \ (2005). High gamma activity in response to deviant auditory stimuli recorded directly from human cortex. {\it Journal of neurophysiology}, {\bf 94}(6): 4269-4280.

[19] Crone, N. E., Korzeniewska, A., \ \& Franaszczuk, P. J. \ (2011). Cortical gamma responses: searching high and low. {\it International Journal of Psychophysiology}, {\bf 79}(1): 9-15.

[20] Flinker, A., Korzeniewska, A., Shestyuk, A. Y., Franaszczuk, P. J., Dronkers, N. F., Knight, R. T., \ \& Crone, N. E. \ (2015). Redefining the role of Broca’s area in speech. {\it Proceedings of the National Academy of Sciences}, {\bf 112}(9): 2871-2875.

[21] Chartier, J., Anumanchipalli, G. K., Johnson, K., \ \& Chang, E. F. \ (2018). Encoding of articulatory kinematic trajectories in human speech sensorimotor cortex. {\it Neuron}, {\bf 98}(5): 1042-1054.

[22] Mesgarani, N., Cheung, C., Johnson, K., \ \& Chang, E. F. \ (2014). Phonetic feature encoding in human superior temporal gyrus. {\it Science}, {\bf 343}(6174): 1006-1010.

[23] Churchland, M. M., Byron, M. Y., Cunningham, J. P., Sugrue, L. P., Cohen, M. R., Corrado, G. S., Newsome, W. T., Clark, A. M., Hosseini, P., Scott, B. B., Bradley, D. C., Smith M. A., Kohn, A., Movshon, J. A., Armstrong, K. M., Moore, T., Chang, S. W., Snyder, L. H., Lisberger, S. G., Priebe, N. J., Finn, I. M., Ferster, D., Ryu, S. I., Sahani, M., \ \& Shenoy, K. V. (2010). Stimulus onset quenches neural variability: a widespread cortical phenomenon. {\it Nature neuroscience}, {\bf 13}(3): 369.

[24] Sun, P., Moses, D. A., \ \& Chang, E. F. \ (2019). Modeling neural dynamics during speech production using a state space variational autoencoder. {\it arXiv:1901.04024.}

[25] Szegedy, C., Ioffe, S., Vanhoucke, V., \ \& Alemi, A. A. \ (2017). Inception-v4, inception-resnet and the impact of residual connections on learning. {\it In Thirty-First AAAI Conference on Artificial Intelligence.}

[26] Mikolov, T., Sutskever, I., Chen, K., Corrado, G. S., \ \& Dean, J. \ (2013). Distributed representations of words and phrases and their compositionality. {\it In Advances in neural information processing systems}, 3111-3119.

[27] Heafield, K., Pouzyrevsky, I., Clark, J. H., \ \& Koehn, P. \ (2013). Scalable modified Kneser-Ney language model estimation. {\it In Proceedings of the 51st Annual Meeting of the Association for Computational Linguistics} {\bf 2}: 690-696.

\end{document}